\documentclass[preprint,12pt]{elsarticle}
\usepackage[a4paper,top=1.5cm,bottom=2cm,left=2cm,right=2cm]{geometry}

\usepackage{amssymb}
\usepackage{amsmath}
\usepackage{lineno}

\usepackage{xcolor}
\usepackage{comment}
\usepackage{caption}
\usepackage{setspace}
\usepackage[colorlinks,linkcolor=blue,bookmarksopen,bookmarksnumbered,citecolor=blue,urlcolor=blue]{hyperref}


\usepackage[shortcuts]{extdash}

\newcommand{\U}[1]{\ensuremath{\mathrm{#1}}}

\journal{Building and Environment}

\begin{document}

\begin{frontmatter}



\title{Inpainting U-Net for seamless pedestrian-level wind prediction across urban morphologies}

\author[aff1]{Jingzi Huang\corref{cor1}}
\author[aff2]{Claire E. Heaney}
\author[aff1]{Tao Li}
\author[aff1]{Xinzhe Li}
\author[aff1]{Graham O. Hughes}
\author[aff1]{Maarten van Reeuwijk}

\cortext[cor1]{Corresponding author. Email address: jingzi.huang17@imperial.ac.uk}

\affiliation[aff1]{organization={Department of Civil and Environmental Engineering, Imperial College London},
            city={London},
            postcode={SW7 2AZ}, 
            country={UK}}

\affiliation[aff2]{organization={Department of Earth Science and Engineering, Applied Modelling and Computation Group, Imperial College London},
            city={London},
            postcode={SW7 2AZ}, 
            country={UK}}

\begin{abstract}
Pedestrian-level wind prediction is essential for urban design and wind-comfort assessment, but high-fidelity simulations such as LES remain computationally expensive, preventing them from being used for rapid evaluation. This study develops a two-stage U-Net framework for efficient prediction of time-averaged pedestrian-level wind speed over realistic urban morphologies. The model is trained and evaluated using the UrbanTALES dataset, which contains realistic city configurations under different approaching wind directions. In the first stage, a baseline U-Net model (M1) predicts the wind field patch-by-patch from normalised building height and fetch information. This patch-wise formulation allows the framework to be applied to urban domains of arbitrary size, but it can introduce discontinuities at patch boundaries because each patch is inferred independently, while limited attention has been paid to assembling such patch-wise predictions into seamless full-field predictions. To address this, a second U-Net model (M2) is introduced as an inpainting-based refinement model. M2 takes a larger contextual window containing the initial M1 prediction and local morphology, and reduces discontinuities in the M1 prediction using neighbouring flow information, resulting in a seamless prediction. During full-field inference, M2 is applied iteratively using a Gauss-Seidel scheme until convergence.
The results show that M1 captures the main spatial distribution of pedestrian-level wind speed and performs well for low- and moderate-velocity regions, although high-velocity peaks are less accurate. M2 substantially reduces patch-boundary artefacts and improves the spatial coherence of the full-field prediction. Across unseen urban cases, the proposed framework reproduces the mean velocity and spatial variability reasonably well, while maximum velocities remain underestimated. Overall, the proposed framework provides an efficient and flexible surrogate model for high-resolution pedestrian-level wind prediction across realistic urban morphologies.

\end{abstract}


\end{frontmatter}


\section{Introduction}
Pedestrian-level wind comfort is a critical consideration in urban planning and design. Wind conditions in cities directly influence natural ventilation, pollutant dispersion, and outdoor thermal comfort \citep{Blocken2016, Tominaga2011}. In particular, excessively strong winds at street level can compromise pedestrian safety and cause significant wind nuisance, especially in the vicinity of tall buildings where high-altitude winds are deflected to the ground \citep{Blocken2004, Stathopoulos2006}. Situated at the base of the atmospheric boundary layer, pedestrian wind is highly sensitive to urban morphological configuration, including building geometry, height variability, and urban density \citep{Blocken2016, He2019}. These factors produce strong spatial heterogeneity in the wind field at the neighbourhood scale ($O(100 \U m)$) \citep{Lean2024}. Therefore, accurate and efficient high-resolution prediction of pedestrian wind is essential for assessing wind comfort and safety in urban environments \citep{Blocken2016}.

Traditional approaches to assessing pedestrian-level wind conditions primarily rely on wind tunnel experiments and computational fluid dynamics (CFD). Wind tunnel testing has long served as the industry standard for evaluating wind comfort and safety in the built environment, particularly in the design of high-rise buildings and complex urban developments \citep{Stathopoulos2006, Blocken2004}. In a boundary-layer wind tunnel, physically scaled models of the target urban morphology are constructed and subjected to a simulated atmospheric boundary-layer flow, with wind speed measurements recorded at pedestrian height using techniques such as hot-wire anemometry, omnidirectional pressure sensors, or particle image velocimetry \citep{Willemsen2007}. However, wind tunnel experiments are inherently limited in scope: constructing physical scale models is costly and time-consuming, the experimental domain is fixed and cannot readily represent different cities or urban configurations, and thermal effects are generally neglected, restricting their applicability for broad or operational pedestrian wind assessment.
High-resolution CFD simulations, particularly Reynolds-Averaged Navier-Stokes (RANS) and large-eddy simulation (LES) approaches, model or partially resolve turbulence and are capable of capturing detailed features of pedestrian-level wind. As a result, both LES and RANS have been widely applied in studying and predicting urban meteorology. However, these CFD methods require substantial computational effort and time, which limits their practicality where rapid pedestrian-level wind predictions are required, such as real-time forecasting or early-stage urban design, where multiple layout alternatives need to be evaluated. Thus, there remains a critical need for approaches that can deliver spatially detailed pedestrian-level wind predictions efficiently and across a wide range of city morphologies.

Data-driven surrogate models have emerged as a promising approach for pedestrian-level wind prediction, offering substantially lower computational cost and faster inference compared to traditional methods \citep{Kastner2023, Gur2024, Wang2025}. Typically, deep learning models are trained on large datasets generated from high-fidelity CFD simulations, enabling them to capture complex nonlinear relationships between urban morphology and microclimate variables; see \citet{Caron2025} for a review. 

Among data-driven approaches, convolutional neural networks (CNNs) have been widely adopted because of their ability to capture the complex, multi-scale spatial structures present in urban air flows. Early studies of urban flows typically adopted a two-stage approach in which a convolutional autoencoder (a particular type of CNN) was used to compress variables such as velocity fields from higher-dimensional physical space to a lower-dimensional latent space and then a second neural network was trained to learn the evolution of the latent variables in time~\cite{Xiang2021,Masoumi-Verki2022,Quilodran-Casas2023}. Although these studies demonstrate the potential of surrogate models, limitations include degradation of the quality of prediction over long time periods and an inability of the models to generalise to unseen building configurations. Recently, the U-Net, a type of CNN originally developed for biomedical image segmentation \citep{Unet2015}, has become the dominant neural-network architecture for pedestrian-level wind prediction \citep{Lu2022, Clarke2024, Cui2025, Briegel2025, Lu2025, Vargiemezis2025, Snaiki2026}. It compresses the input into increasingly compact representations with high-level flow structures, then progressively upsamples to reconstruct the full-resolution output field. Its defining feature is the skip connections between encoder and decoder layers at each resolution level, allowing the model to retain high-resolution spatial information across scales and reconstruct detailed output fields. This multi-scale hierarchical representation makes the U-Net particularly well-suited for urban wind emulation, where accurate prediction requires the simultaneous representation of large-scale channelling and small-scale corner accelerations.

To address computational limitations, the U-Net was originally proposed with a patch-based training strategy, in which the input domain is subdivided into tiles of a fixed size that are processed independently. This strategy is well established across spatial prediction tasks, including medical image segmentation \citep{Unet2015}, remote sensing \citep{Maggiori2017}, and aerial scene parsing \citep{Volpi2017}. 
Similar patch-based formulations have also been developed for wind-field prediction. A number of studies extract training data from patches taken from larger domains \citep{Clemente2023, Lu2025, Briegel2025, Calafell2026} and make predictions for patches with unseen building configurations. However, limited attention has been given to assembling such patch-wise predictions into coherent full-field predictions of large urban areas. 

A fundamental challenge in stitching patch-wise predictions together to form a full-field prediction is inconsistency at patch boundaries: because each patch is predicted independently without knowledge of its neighbours, discontinuities arise at patch boundaries when the results are assembled into a global solution~\citep{Reina2020, Innamorati2020}. Although such inconsistencies can be mitigated with post-processing, including overlapping-window averaging and weighted blending \citep{Unet2015, Isensee2018,Pielawski2020, Abdellatif2024, Jeon2025}, these methods smooth discontinuities statistically without incorporating surrounding information and have no explicit physical basis. Another approach to removing discontinuities at patch boundaries is to constrain the neural-network prediction of a patch by using predictions from the surrounding patches and iterating~\citep{Heaney2022}. Deep-learning--based inpainting methods~\citep{Pathak2016, Liu2018, Yu2018_inpainting} explicitly exploit information from surrounding regions when reconstructing missing areas. In the context of urban flow prediction, spatial context from the surrounding neighbourhood can be used to guide the prediction of a neural network in a manner similar to the influence of boundary conditions on a flow. 

Thus, this study aims to develop an inpainting-based framework that stitches patch-wise predictions into a seamless full-field prediction with no discontinuities at patch boundaries. Specifically, we propose a two-stage U-Net framework for predicting time-averaged pedestrian\-/level wind speed from urban topography and user-specified upstream wind conditions. In the first stage, the baseline model (M1) performs patch-wise inference over the full domain, dividing the urban morphology into non-overlapping patches and assembling the results into an initial full-field wind estimate. In the second stage, the inpainting model (M2) refines the full-field prediction produced by M1 using a larger patch. The central region of each M2 patch corresponds to the original M1 patch size, while the surrounding margins provide additional neighbourhood context for reducing boundary discontinuities and improving local coherence. 
Together, these models enable high-resolution prediction of pedestrian-level wind fields across urban domains of arbitrary size. The workflow and methodology are described in detail in \S~\ref{sec: methodology}. The results and model performance are presented in \S~\ref{sec: results}, and the conclusions are outlined in \S~\ref{sec: conclusions}.

\section{Methodology} \label{sec: methodology}
This section first introduces the dataset used to train the models, and then describes the two-stage prediction framework consisting of a baseline model (M1) and a refinement model (M2) in \S~\ref{sec: two-stage framework}. The overall framework operates with normalised variables, and the final prediction is mapped back to the physical values only after the full-field inference procedure is completed. Thus, the data processing methods are introduced in \S~\ref{sec: data_preprocessing}.

\subsection{UrbanTALES dataset}

The models are trained and evaluated using a subset of the UrbanTALES (Urban Turbulent Airflow using systematic Large Eddy Simulations) dataset \citep{Nazarian2025}, a large-scale LES database designed for studying turbulent airflow over urban environments. UrbanTALES provides high-fidelity, time-averaged flow fields generated using the Parallelised Large-Eddy Simulation Model (PALM, a widely used LES software \citep{PALM2015}), together with the corresponding urban morphology information. The full dataset contains both idealised building-array cases and realistic urban-neighbourhood cases, covering a broad range of urban densities, building layouts, height distributions, and approaching wind directions.

In this study, we use the realistic urban neighbourhood component of UrbanTALES, which contains urban geometries sampled from major cities worldwide. Specifically, 288 realistic urban configurations are selected for model training and evaluation. These configurations represent a wide variety of urban forms, from sparse low-density neighbourhoods to compact high-density city blocks, thereby providing diverse morphological conditions for learning the relationship between urban geometry and pedestrian-level wind speed. The inclusion of multiple approaching wind directions, including cases such as $0^\circ$, $45^\circ$, and $90^\circ$, further increases the variability of the flow patterns and enables the models to learn direction-dependent effects associated with building arrangement and upstream exposure.

All LES cases used in this study are provided on a uniform horizontal grid resolution of $1$ m. The pedestrian-level wind speed is extracted at a height of $1.75$ m above ground, corresponding to the typical height used to assess pedestrian wind conditions. 

\subsection{Training and inference of two-stage framework} \label{sec: two-stage framework}
\begin{figure*}
    \centering
    \includegraphics[width=16cm]{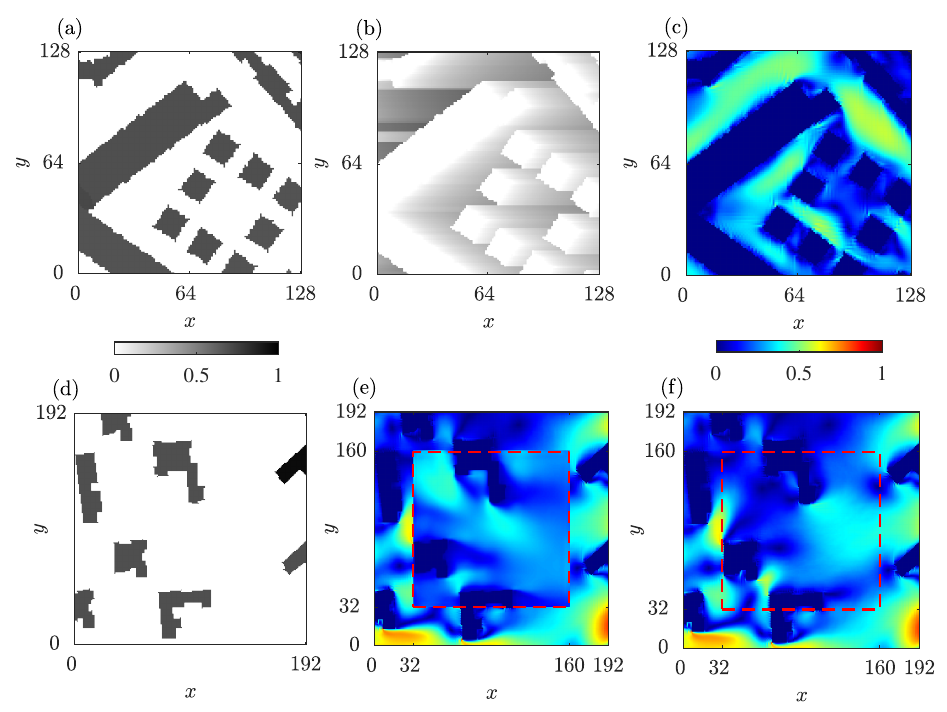}
    \caption{Examples of extracted patches used for model training. (a-c) Patches used in M1: (a) input topography, (b) input fetch, and (c) the corresponding ground-truth velocity field at pedestrian level. (d-e) Patches used in M2: (d) input topography, and (e) input speed field, where the central region (indicated by dashed lines) is predicted by M1, while the surrounding margins are taken from the ground truth. (f) Ground-truth velocity field in the central region of (e). All the fields are normalised.}
    \label{fig:Fig1}
\end{figure*}

The framework aims to predict the time-averaged pedestrian-level wind speed for a given urban morphology under prescribed upstream wind conditions. The input information consists of the urban morphology, represented by building locations and heights, and a reference upstream wind velocity. This reference velocity is defined at a height of $z = 30$ m, including both its magnitude (denoted as $U_{30}$) and direction. The use of a reference wind velocity at a fixed height is standard practice in wind engineering \citep{Holmes2015} and is obtainable from meteorological observations or numerical weather prediction models \citep{Stull1988}.

\subsubsection{Training of the baseline model (M1)}
The baseline model M1 learns the primary mapping from urban morphology to pedestrian-level wind speed at a patch scale of $128\, \U m\times 128\, \U m$ (corresponding to $128 \times 128$ pixels as the LES resolution is $1$ m). Specifically, M1 is trained using patches with two input channels: normalised topography (a building mask incorporating height information, e.g., Fig.~\ref{fig:Fig1}a) and the normalised fetch under the given wind direction (Fig.~\ref{fig:Fig1}b), both derived from morphological information. Details of preprocessing, including patch generation, normalisation and fetch calculation, are provided in \S~\ref{sec: data_preprocessing}. The target output is the corresponding normalised ground-truth wind speed within the same patch (Fig.~\ref{fig:Fig1}c). 

\subsubsection{Training of the refinement model (M2)}
The refinement model M2 is trained to refine the local prediction produced by M1 using information from the surrounding flow environment. It operates on a larger patch size of $192\, \U m \times 192\, \U m$, enabling a broader spatial scope. Each M2 input sample contains two channels. The first channel is the normalised building height over the $192\, \U m \times 192\, \U m$ patch centred on the original $128\, \U m \times 128\, \U m$ patch of interest (Fig.~\ref{fig:Fig1}d). The second channel is a composite normalised wind-speed field, in which the central $128^2$ region is filled with the corresponding M1 prediction for that patch, while the surrounding margins are taken from the neighbouring ground-truth flow field (Fig.~\ref{fig:Fig1}e). The target output of M2 is the ground-truth wind-speed field of this $192^2$ patch (Fig.~\ref{fig:Fig1}f). In this way, M2 learns to correct local errors and boundary inconsistencies in the M1 estimate according to the surrounding flow context.
 

\subsubsection{Two-stage full-field inference}
During full-field inference, the original urban domain is first periodically tiled into a $3 \times 3$ extended field, with the target domain located at the centre tile. This ensures that the models work in the field after rotation. The wind direction, defined as the angle from the positive horizontal $x$-axis to the wind direction, with positive angles measured anticlockwise, is not explicitly provided as an input to the models. Both models are trained under the assumption that the streamwise direction aligns with the positive $x$-axis. Therefore, before inference, all fields are rotated such that the streamwise direction coincides with the positive $x$-axis. The procedures are introduced in \S~\ref{sec: expansion and rotation}. M1 is then applied to the rotated extended field using non-overlapping $128^2$ patches. For each patch, the normalised building height and normalised upstream fetch are used as input channels, and the patch-wise predictions are assembled to form an initial normalised estimate of the pedestrian-level wind field. M2 is then applied to refine this initial estimate. For each M2 patch, its central $128^2$ region coincides with the corresponding M1 patch, and the model takes the local $192^2$ building-height field and the current estimated $192^2$ wind-speed field as input. After M2 inference, the estimated wind-speed field is updated using the modified wind-speed field in each $192^2$ patch.

It should be noted that, unlike in the M2 training stage, where the surrounding margins are provided by ground-truth flow fields, no ground-truth flow information is available during full-field inference. The surrounding margins are taken from the initial M1 prediction and are subsequently updated by M2. To improve consistency between neighbouring patches, M2 is applied using an under-relaxed Gauss-Seidel iterative scheme. During each iteration, selected patches are processed sequentially. The corresponding field values are immediately updated and become available as context for subsequent patch predictions within the same iteration. For each pixel $i$ in the updated central region, the local under-relaxed update is defined as
\begin{equation}
    \hat{U}^{\mathrm{new}}_i
    =
    (1-\alpha)\hat{U}^{\mathrm{old}}_i
    +
    \alpha \hat{U}^{\mathrm{M2}}_i ,
\end{equation}
where $\hat{U}^{\mathrm{old}}_i$ is the speed field value immediately before the local patch update, $\hat{U}^{\mathrm{M2}}_i$ is the direct speed prediction from M2, $\hat{U}^{\mathrm{new}}_i$ is the relaxed value written back to the field, and $\alpha=0.2$ is the relaxation factor.
The convergence of the iterative refinement is monitored using the mean absolute difference between two consecutive iterations over the valid pixels in the M2 update region:
\begin{equation}
    D^{(k)}
    =
    \frac{1}{N_{\Omega_{\mathrm{M2}}}}
    \sum_{i \in \Omega_{\mathrm{M2}}}
    \left|
    \hat{U}^{(k)}_i
    -
    \hat{U}^{(k-1)}_i
    \right|,
\end{equation}
where $\Omega_{\mathrm{M2}}$ denotes the set of finite pixels within the region updated by M2, $N_{\Omega_{\mathrm{M2}}}$ is the number of pixels in $\Omega_{\mathrm{M2}}$, and $\hat{U}^{(k)}_i$ and $\hat{U}^{(k-1)}_i$ are the field values at the end of iterations $k$ and $k-1$, respectively. The iteration is stopped when $D^{(k)}$ fails to improve for three consecutive iterations, or when the prescribed maximum number of iterations, set to 50 in this study, is reached. The final M2 prediction is then taken from the iteration that gives the minimum value of $D^{(k)}$.

Since M2 uses the predicted field as its neighbourhood context during inference, the final result remains dependent on the quality of the initial M1 prediction. Errors in the M1 estimate may propagate or accumulate during the iterative inpainting process. To reduce unnecessary inference and limit error propagation outside the target area, M2 is applied only to patches located near the rotated centre tile corresponding to the original urban domain, rather than to the entire $3 \times 3$ extended field. In practice, this update region is obtained by expanding the original urban-domain mask outward by $32$ pixels, ensuring that all pixels in the original domain can be covered by the central $128^2$ region of at least one M2 patch.

After the iterative refinement is completed, the refined field corresponding to the original urban domain is cropped from the rotated extended prediction, inverse-rotated to the original orientation, and denormalised to recover the physical wind-speed values.

\subsection{Data preprocessing} \label{sec: data_preprocessing}
To serve as training inputs and targets, the LES dataset is normalised, rotated, and cropped into patches of fixed size. This subsection describes the preprocessing workflow that transforms the original dataset into the patches shown in Fig.~\ref{fig:Fig1}. The whole procedure is sketched in Fig.~\ref{fig:Fig2}.

\subsubsection{Normalisation}
To ensure numerical stability and facilitate efficient gradient-based optimisation, the dataset values are constrained to lie in the unit interval. The topography data from the LES is represented as a 2-D matrix of building heights $H(x,y)$ (e.g., Fig.~\ref{fig:Fig2}a), where $x$ and $y$ are horizontal and vertical axes from the plan view, respectively, and the absence of a building corresponds to $H = 0$. The topography is normalised using the $95^\mathrm{th}$ percentile of the entire training set:
\begin{equation}
    h(x, y) = \frac{\ln (1 + H)}{\ln (1 + H_{95\%})} \, .
\end{equation}
Most values of $h$ lie within $[0,1]$, and any values outside this range are clipped to $[0,1]$.

The upstream fetch is also extracted as an input for M1. The fetch $L$ is defined as the distance from a given point to its nearest upwind building along the approaching wind direction. This distance is normalised by the upstream building height $H_u$ to obtain the ratio $r = L/H_u$, which is expected to characterise the shielding effect --- a key factor influencing the flow velocity at the position of interest. The fetch ratio is then normalised in the same manner as the topography based on the entire training dataset:
\begin{equation}
    l(x, y) = \frac{\ln (1 + r)}{\ln (1 + r_{95\%})} \, .
\end{equation}
By construction, locations farther from the upwind building tend to have $l$ approaching $1$, whereas locations just downstream of a building tend to have $l$ near $0$. The value of $l$ at building locations is defined to be $0$ (e.g., see Fig.~\ref{fig:Fig1}b).

The 2-D wind speed field at pedestrian level is computed as $U = \sqrt{\overline{u}^2 + \overline{v}^2}$ (e.g., Fig.~\ref{fig:Fig2}b), where $\overline{u}$ and $\overline{v}$ are the time-averaged horizontal velocity components. The wind speed field $U$ is first non-dimensionalised by its corresponding reference velocity $U_{30}$ in each case (the wind speed at $z = 30$ m), yielding $\hat{u} \equiv U/U_{30}$. Subsequently, $\hat{u}$ is rescaled by its $99^\mathrm{th}$ percentile value over the entire training set:
\begin{equation} \label{eq: f}
    f(x, y) = \frac{\hat{u}}{\hat{u}_{99\%}} \, .
\end{equation}
An example of the normalised velocity field is shown in Fig.~\ref{fig:Fig2}(c). Note that the raw LES velocity field contains NaN values at building locations; these values are set to $0$ prior to normalisation and are excluded when computing the $99^\mathrm{th}$ percentile for normalisation purposes. The value of $\hat{u}_{99\%}$ is used to recover the physical velocity at the end of the workflow.
 
\subsubsection{Expansion and rotation} \label{sec: expansion and rotation}
\begin{figure*}
    \centering
    \includegraphics[width=16cm]{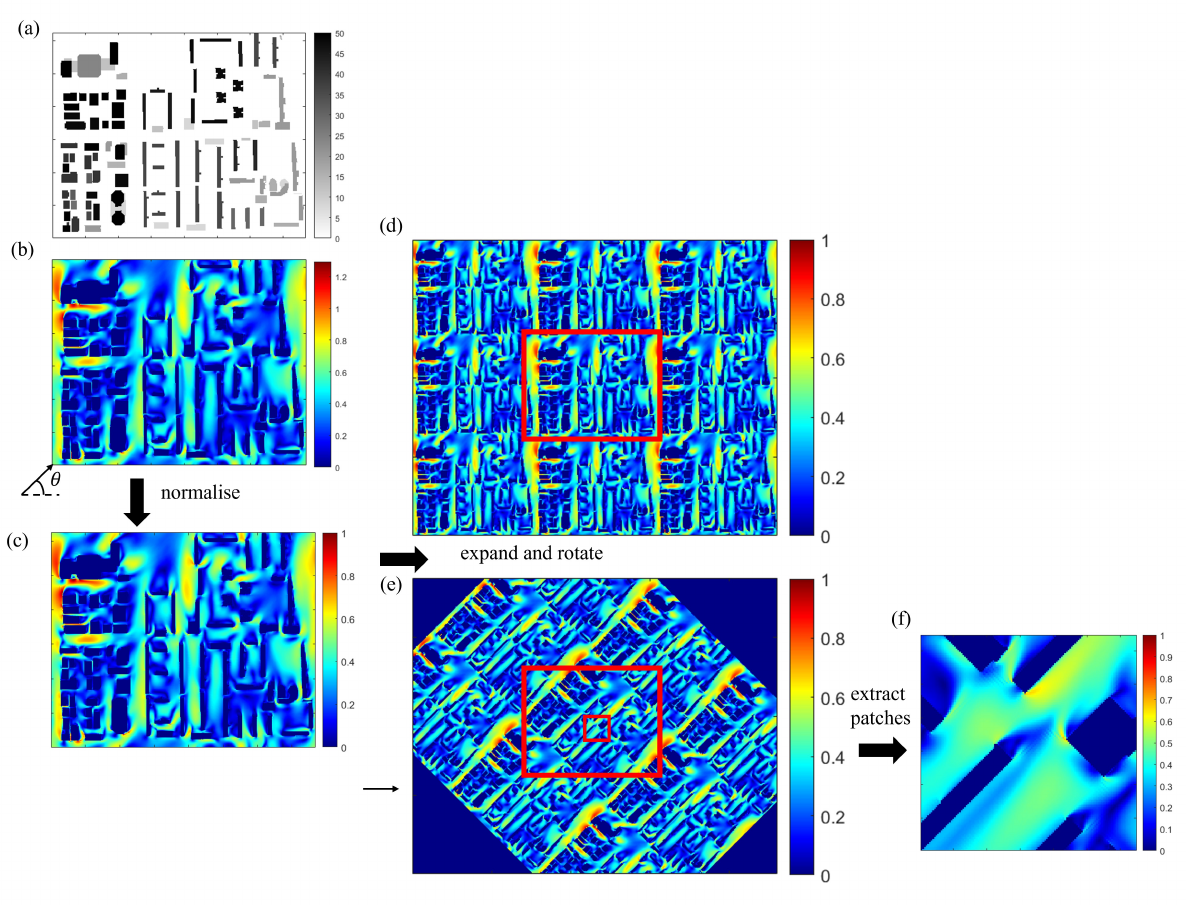}
    \caption{Dataset preprocessing illustrated using the wind speed field as an example. (a, b) The original LES topography and wind speed fields, respectively, with the upstream wind direction $\theta$. (c) The normalised wind speed field. (d) Expansion of (c) into a $3 \times 3$ tiling under periodic boundary conditions. The central red rectangle indicates the location of the original field. (e) The clockwise rotation of (d) by the angle $\theta$, aligning the streamwise direction with the positive $x$-axis. The large red rectangle marks the same region as in (d). (f) An example patch extracted from the small red rectangle in (e).}
    \label{fig:Fig2}
\end{figure*}

As shown in Fig.~\ref{fig:Fig2}(d), the original normalised fields (the area marked by a red rectangle) are extended into a $3 \times 3$ tile under periodic boundary conditions. 
This periodic expansion allows patches near the original field boundaries to be extracted without out-of-domain issues. Note that UrbanTALES applies periodic boundary conditions on lateral walls.

Then the entire field is rotated by the inverse wind direction $\theta$ (shown in Fig.~\ref{fig:Fig2}e), such that the approaching wind is aligned with the positive $x$-axis. During inference, the input fields are first rotated to align the streamwise direction with the $x$-axis. Prediction is then performed along this direction, and the resulting output fields are rotated back to their original orientation once inference is completed.

\subsubsection{Patch extraction}
The patches are extracted from the expanded and rotated normalised fields by randomly locating their centres (e.g., Fig.~\ref{fig:Fig2}f). The patch centres are randomly sampled but confined to the region of the original field prior to expansion and rotation, i.e., within the large red rectangle in Fig.~\ref{fig:Fig2}(e).
Although the extraction is random, two measures are applied to ensure the patches are informative and not overly redundant. First, the number of patches per case is limited to
\begin{equation}
    N = 3 \times \frac{A}{A_{patch}} \, ,
\end{equation}
where $A$ is the area of the original field and $A_{patch}$ is the area of the patch. Second, the patch centres are required to be at least half a patch size apart to reduce overlap. These requirements ensure that the model learns diverse local features while maintaining consistent coverage of the original domain.

Among the 288 LES cases used in this study, 18 cases covering different urban morphologies and approaching wind directions are reserved as completely unseen cases for full-field prediction tests. These cases are excluded from all training and validation procedures. The remaining 270 cases are then divided at the case level into training and validation sets, with $80\%$ used for training and $20\%$ used for validation. This case-level separation ensures that patches extracted from the same LES case do not appear in both the training and validation sets. Finally, approximately $11 \, 000$ patches are extracted for M1. Note that Fig.~\ref{fig:Fig2} illustrates the operations on the velocity field for M1 as an example. The same training and validation partition is used to generate the M2 patches, but with two differences: the patch size is increased, and the central region of the training velocity field is replaced by the corresponding prediction from M1.

\subsection{U-Net architecture and training configuration}
\begin{figure*}
    \centering
    \includegraphics[width=15cm]{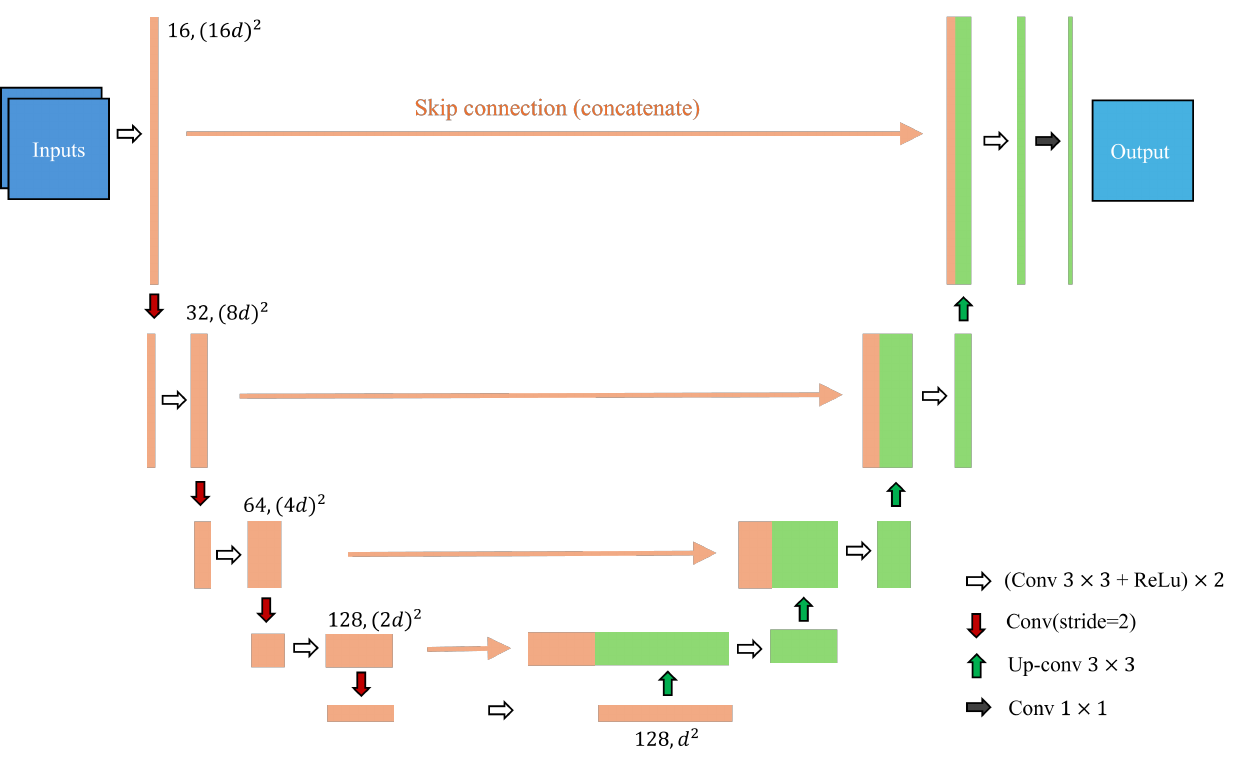}
    \caption{Schematic diagram of the U-Net architecture used in this study. Both models share the same architecture, with their unique inputs and the output shown in Fig.~\ref{fig:Fig1}. The dimension of the features at each stage is labelled. For M1, the input patch dimension is $(16d)^2 = 128^2$, while for M2, it is $(16d)^2 = 192^2$. Accordingly, the spatial dimension of the feature map after the downsampling (encoder) stage is $d^2$. The number before the patch dimension is the number of channels.}
    \label{fig:Fig3}
\end{figure*}

The U-Net architecture design follows the original architecture of \citet{Unet2015}, which has demonstrated strong performance in image-to-image regression tasks. Both M1 and M2 employ the same architecture, adopting a fully convolutional encoder-decoder structure. The encoder functions as a multiscale feature extractor, learning geometric and flow-related representations at gradually coarser resolutions, while the decoder reconstructs the target wind speed field by combining high-level latent features with fine-scale information transmitted through skip connections.

The proposed U-Net consists of four levels of resolution (see the schematic diagram of the architecture in Fig.~\ref{fig:Fig3}). At the input stage, the data is first projected onto 16 feature channels through a double-convolution block, i.e. two consecutive $3 \times 3$ convolutional layers, each followed by a ReLU activation. 
In the encoder, spatial resolution is reduced progressively through four downsampling stages. Each stage first applies a stride convolution to halve the spatial resolution, followed by a double convolution to extract features and increase the channel dimension. The number of channels grows from 16 to 32, 64, and 128, and remains 128 at the deepest level. A dropout rate of 0.05 is applied in the deepest layer, where $5\%$ of the activations are randomly set to zero during training to avoid overfitting and improve regularisation.

In the decoder, feature maps are upsampled using bilinear interpolation followed by a convolution, and then concatenated with the corresponding encoder features via skip connections. Each concatenated feature map is processed by a double convolution to recover spatial resolution and reduce the channel dimension.

Although M1 and M2 share the same architecture, the sizes of their inputs and targets are different. 
M1 takes two $128 \times 128$ input channels (normalised topography and fetch) and predicts a $128 \times 128$ normalised velocity field. M2 takes two $192 \times 192$ input channels (normalised topography and velocity) and predicts a $192 \times 192$ normalised velocity field.
In both models, ReLU activation functions are used in all the hidden layers and a sigmoid activation is applied at the final $1 \times 1$ convolution layer to constrain the normalised prediction to the range (0,1). The models are trained using the AdamW optimiser for up to 1000 epochs, with a learning rate of $10^{-4}$ and a mean absolute error (MAE) loss function, on an AMD MI210 64GB GPU.

\subsection{Performance evaluation metrics}
Different urban applications place emphasis on different characteristics of the wind field. For example, pedestrian wind comfort and building ventilation studies often focus on the accuracy of mean wind speed, while pollutant dispersion and urban climate studies are more sensitive to the spatial distribution and flow structures. In addition, risk-related applications (e.g., extreme wind assessment) require a reliable prediction of high-magnitude events. Therefore, the performance of the models is evaluated using three complementary metrics. For consistency, let $U_i$ and $\hat{U}_i$ denote the ground-truth and predicted wind speed at pixel $i$, respectively, and let $\Omega$ denote the set of pixels over which a metric is evaluated, with $N_{\Omega}$ being the number of pixels in $\Omega$. The coefficient of determination, $R^2$, measures the proportion of variance in the reference data explained by the model predictions,
\begin{equation}
    R^2
    =
    1 -
    \frac{\sum_{i \in \Omega} (U_i - \hat{U}_i)^2}
    {\sum_{i \in \Omega} (U_i - \overline{U}_{\Omega})^2},
\end{equation}
where $\overline{U}_{\Omega}=N_{\Omega}^{-1}\sum_{i \in \Omega} U_i$ is the mean reference value over $\Omega$. $R^2$ primarily evaluates the agreement in overall magnitude and large-scale trends of the wind field. 

The normalised mean absolute error, NMAE, quantifies the average magnitude of prediction errors relative to a characteristic scale of the data,
\begin{equation}
    \mathrm{NMAE}
    =
    \frac{1}{N_{\Omega}}
    \sum_{i \in \Omega}
    \frac{|U_i - \hat{U}_i|}{U_{\mathrm{ref}}},
\end{equation}
where $U_{\mathrm{ref}}=\max_{i \in \Omega}(U_i)-\min_{i \in \Omega}(U_i)$ is the reference range over the evaluated pixels. Compared to $R^2$, NMAE provides a more direct and interpretable measure of absolute error and is less sensitive to outliers. Note that 
\begin{equation}
\mathcal{L}_{\mathrm{MAE}} =
\frac{1}{N_{\Omega}} \sum_{i \in \Omega}
\left| U_i - \hat{U}_i \right| \, ,
\end{equation}
is the MAE loss function applied.

Finally, the structural similarity index, SSIM \citep{Wang2004}, was originally developed for image quality assessment and evaluates the similarity between two fields in terms of luminance, contrast, and structure. The score is computed locally: a Gaussian-weighted window is slid across the two fields, and at each position the local statistics of the reference and predicted patches, denoted by $\mathbf{U}$ and $\hat{\mathbf{U}}$, are compared. Specifically, $\mu_{\mathbf{U}}$ and $\mu_{\hat{\mathbf{U}}}$ are the Gaussian-weighted local means, $\sigma_{\mathbf{U}}^2$ and $\sigma_{\hat{\mathbf{U}}}^2$ are the corresponding local variances, and $\sigma_{\mathbf{U}\hat{\mathbf{U}}}$ is the local covariance between the two patches. The local SSIM is then
\begin{equation}
    \mathrm{SSIM}(\mathbf{U}, \hat{\mathbf{U}})
    =
    \frac{
    (2\mu_{\mathbf{U}}\mu_{\hat{\mathbf{U}}} + C_1)
    (2\sigma_{\mathbf{U}\hat{\mathbf{U}}} + C_2)
    }
    {
    (\mu_{\mathbf{U}}^2 + \mu_{\hat{\mathbf{U}}}^2 + C_1)
    (\sigma_{\mathbf{U}}^2 + \sigma_{\hat{\mathbf{U}}}^2 + C_2)
    },
\end{equation}
where $C_1 = (k_1 L)^2$ and $C_2 = (k_2 L)^2$ are small constants that stabilise the division when the local means or variances are near zero, with $L$ the dynamic range of the field values and $k_1 = 0.01$, $k_2 = 0.03$ following \citet{Wang2004}. The overall SSIM score is the spatial mean of the local values over all windows. SSIM values range from $-1$ to $1$, with $1$ indicating perfect structural agreement. Unlike the two point-wise error metrics above, SSIM emphasises spatial organisation and is therefore particularly suitable for assessing the preservation of flow patterns and coherent structures in the predicted wind field.

\section{Results} \label{sec: results}
The predictive performance of the proposed models on unseen samples is assessed in this section, following the workflow described in \S~\ref{sec: two-stage framework}.

\subsection{Patch-level performance of M1}
\begin{figure*}
    \centering
    \includegraphics{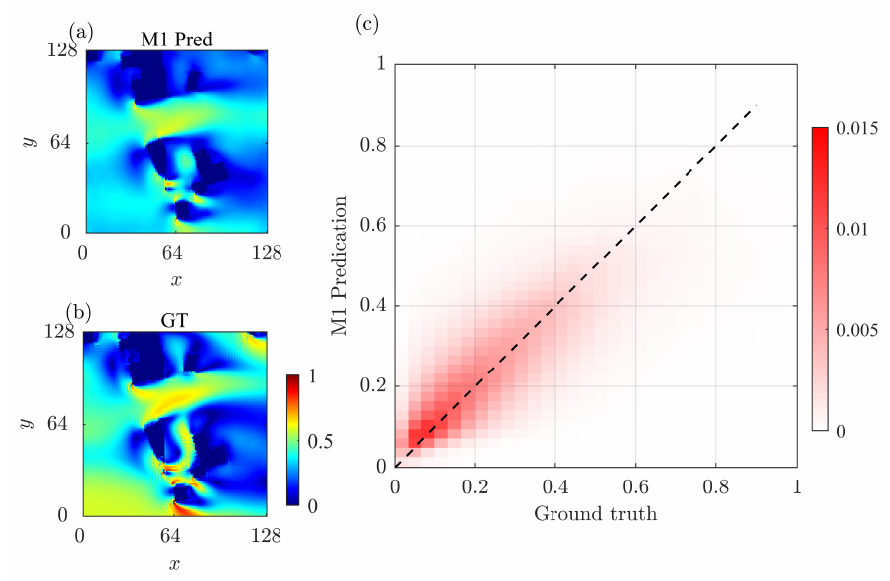}
    \caption{Example M1 prediction for an unseen patch: (a) predicted wind speed, (b) corresponding ground truth. (c) Probability density function (PDF) comparing predicted and true pixel velocities across all unseen test patches. The dashed line indicates perfect agreement. The wind speeds shown here are normalised.}
    \label{fig:Fig4}
\end{figure*}
\begin{figure*}
    \centering
    \includegraphics{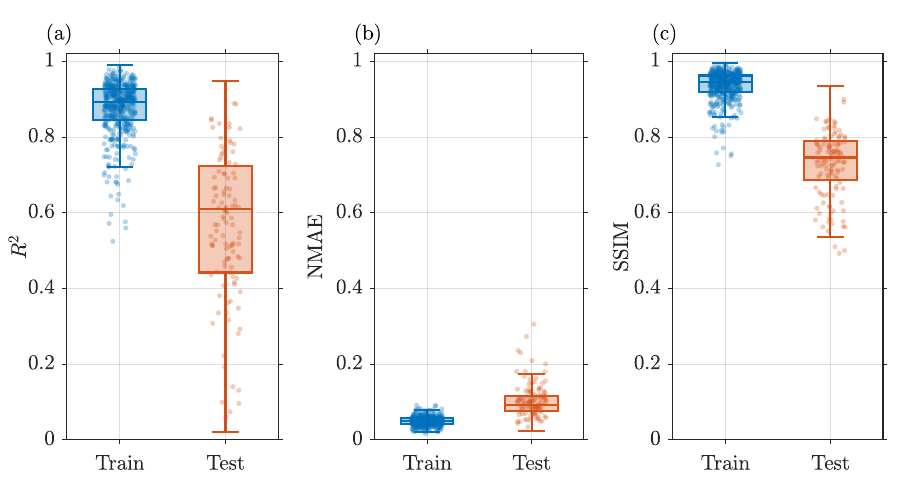}
    \caption{Performance of M1 on training and test data patches, shown across three metrics.}
    \label{fig:Fig5}
\end{figure*}

Figure~\ref{fig:Fig4}(a) shows an example of predictions of normalised wind speeds by model M1 on unseen samples, compared with the ground truth (GT) in Fig.~\ref{fig:Fig4}(b). The model captures the majority of low-velocity regions well, including areas occupied by buildings where the velocities are expected to be very close to zero. Notably, in the training input and ground truth, velocities within building regions are strictly set to zero, while M1 predicts these regions with low velocities of the order of $10^{-5}$. Although building areas can eventually be identified using the provided topography mask in the full-field prediction, this performance still demonstrates the capability of M1 to predict low velocities.

However, the prediction is not accurate at higher velocities (e.g., above 0.6). In particular, although the model is still able to capture the presence and general shape of the wake region induced at the building corners, it consistently underestimates the peak velocity there. It is worth noting that the velocity is normalised to unity, and the reduced accuracy at high velocities may be related to the scarcity of samples near the upper bound, as well as the tendency of convolutional neural networks to smooth extreme values.

Figure~\ref{fig:Fig4}(c) presents the probability density function (PDF) of predicted and true velocities over all pixels across all unseen testing patches. The scatter points are generally distributed around the line of perfect agreement, indicating an overall good predictive performance. 
The figure also indicates that, due to normalisation, most velocity values are rescaled to below $0.6$. Thus, the high-velocity regime, where M1 performs relatively poorly, does not constitute the majority of cases overall. This also reveals a possible explanation for the weaker performance: samples with values above $0.6$ are relatively rare in the dataset, resulting in limited training data for this regime.

The patch-level performance of M1 is shown in the box plots (Fig.~\ref{fig:Fig5}) for three evaluation metrics ($R^2$, NMAE and SSIM) across all training and test patches. For the training set, all metrics exhibit nearly ideal values with a narrow spread, indicating that the network has effectively learned and fitted the data. Although a few training samples fall below the lower quartile in $R^2$ and SSIM, they are only a small fraction of the total.

The unseen test set allows us to assess the generalisation capabilities of the model. Overall, although the performance is not as strong as for the training set, the model delivers satisfactory predictions for the majority of test patches, with relatively high mean $R^2$ and SSIM and relatively low mean NMAE. Notably, the $R^2$ box plot for the test set is more spread out, with whiskers extending nearly across the full 0 to 1 range. This indicates that, although many predictions are accurate, a substantial portion of patches exhibit relatively low $R^2$. This variability may be attributed to many factors. First, the test set may include flow conditions that are less represented in the training dataset, such as the high-velocity regimes discussed above. Second, even within the training set, there exist samples with only moderate predictive accuracy (e.g., $R^2$ in the range of $0.5-0.7$), indicating challenging cases for the model. When similar or more complex patterns appear in the unseen test data, these limitations can lead to further performance degradation, thereby contributing to the wider spread of $R^2$ values.

\begin{figure*}
    \centering
    \includegraphics{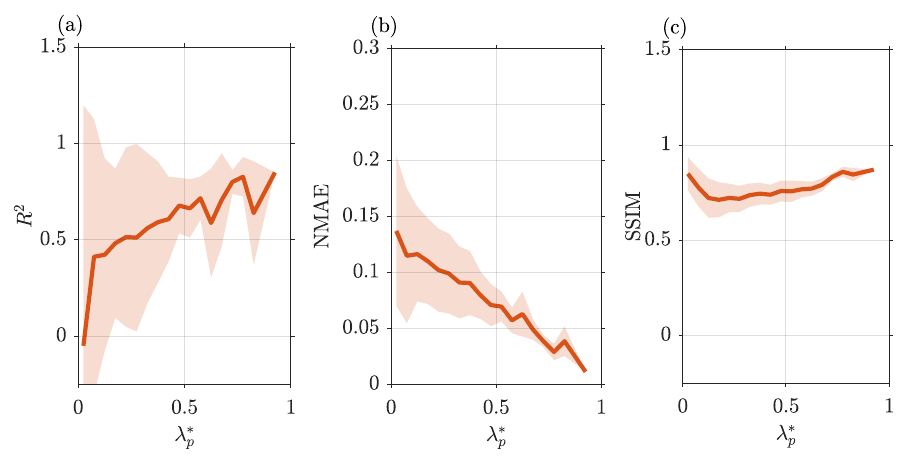}
    \caption{Mean values of three performance metrics for test data patches as a function of the building plane area index of the patches. Shaded bands indicate one standard deviation above and below the mean.}
    \label{fig:Fig6}
\end{figure*}

A key factor influencing the performance of M1 across patches is the building density of the patch, $\lambda_p^*$, defined as the ratio of building footprint area to the total patch area. A challenge inherent to this patch-wise model is the limited spatial perspective, although fetch information can provide some cues from outside the patch. In particular, patches with very low building coverage, which can be extracted from cases with large open spaces, provide little morphological information to the model, making accurate prediction difficult.
Figure~\ref{fig:Fig6} illustrates the relationship between patch building density $\lambda_p^*$ and model performance across the three evaluation metrics. The mean values of $R^2$ and NMAE in Figs.~\ref{fig:Fig6}(a,b) indicate that the model performs poorly on low-density patches, as reflected by both the metric values and their large variance, where only limited building information is available, such as patches dominated by large open spaces. This behaviour may also be due to the smaller number of samples in this density range.
However, performance improves as the building density increases, as indicated by higher $R^2$, lower NMAE, and reduced variance. Note that the metrics are calculated only over the fluid field outside the buildings; therefore, this improvement suggests that greater building coverage provides more morphological information, leading to more accurate predictions. Interestingly, Fig.~\ref{fig:Fig6}(c) shows that SSIM remains relatively stable across $\lambda_p^*$, likely because it captures structural similarity from a perspective distinct from $R^2$ and NMAE.

\subsection{Patch-level performance of M2}

As mentioned, one key challenge when M1 predicts the full field patch-by-patch is handling the inconsistencies that arise at the patch boundaries. However, this is not a fault of M1 itself, because M1 is designed to predict the wind field based on the local morphology of a single patch and is not trained to enforce smoothness across neighbouring patches.
Although Gaussian-weighted blending can be applied to smooth the discontinuities, this would act more like a post-processing stage and would not lead to a physically meaningful solution. Therefore, the second stage of the framework, M2, is introduced to refine the M1 predictions by incorporating information from the surrounding area, thereby avoiding inconsistencies at patch boundaries.

\begin{figure*}
    \centering
    \includegraphics[width=16cm]{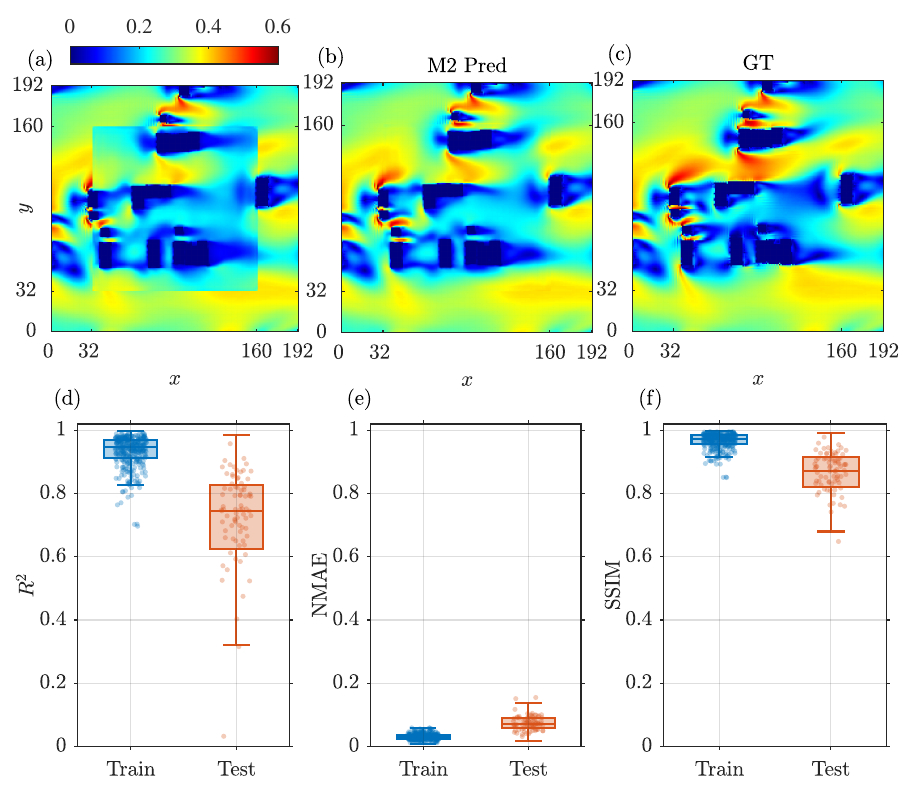}
    \caption{Performance of M2: (a) an example patch of the input wind field for M2 used during training, (b) the prediction of M2 for the patch in (a) and associated morphology (not shown), (c) the corresponding ground truth. (d-f) The three performance metrics. The wind speeds shown here are normalised.}
    \label{fig:Fig7}
\end{figure*}

Figure~\ref{fig:Fig7}(a) presents an example input patch for M2. The patch has a size of $192 \times 192$, where the central $128 \times 128$ region (i.e., from $x,y = 32$ to $160$) corresponds to the normalised wind field predicted by M1 based on the underlying morphology, while the surrounding margin is taken from the ground truth. Figures~\ref{fig:Fig7}(b) and (c) show the output of M2 and the ground truth, respectively.
The more coherent structure in Fig.~\ref{fig:Fig7}(b) shows that M2 not only smooths the discontinuities at the patch boundaries of the M1 predictions, but also refines the results in these regions. Surprisingly, it even improves the prediction of high-velocity regions by using the high-velocity information supplied at the boundary. However, at the patch centre, M2 largely preserves the predictions from M1. 
This indicates that M2 primarily acts near the boundary of the M1 patch. In addition, a known limitation of convolutional architectures in M1 is that predictions near the patch boundary are affected by incomplete spatial context and the use of padding, making them less accurate and introducing artefacts \citep{Innamorati2020}. As a result, M2 mitigates these boundary issues.
It is worth noting that an additional input channel to M2 is the $192 \times 192$ morphology corresponding to this patch, although it is not shown here. Both training and evaluation are still conducted using the normalised variables.

Figures~\ref{fig:Fig7}(d-f) present the performance of M2 across three metrics. Overall, M2 demonstrates a more reliable performance than M1 (Fig.~\ref{fig:Fig5}) on both the training and test sets. For example, the $R^2$ distribution for the training set exhibits fewer outliers beyond the lower whisker, while for the unseen test set, the spread of the whiskers is significantly reduced, indicating more consistent predictions.
However, this result is expected to some extent, as M2 is trained with additional neighbouring ground-truth information, whereas M1 relies solely on morphological information.

\subsection{Full-field prediction performance}

\begin{figure*}
    \centering
    \includegraphics[width = 17cm]{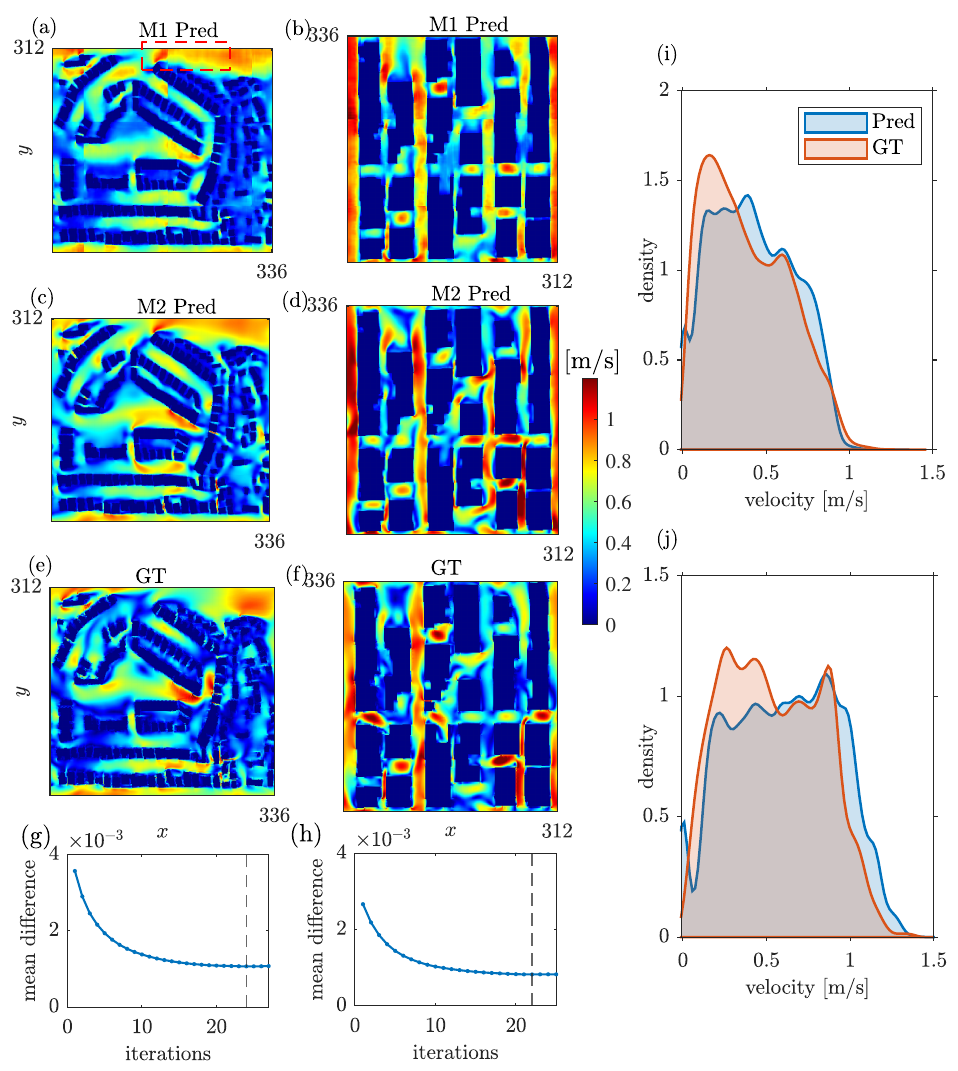}
    \caption{Comparison of full-field wind field predictions between the proposed framework and LES for two representative cases. (a, c, e) Case AU-Syd-U1 with $0^\circ$ prevailing wind direction; (b, d, f) Case CA-Van-U2 with $90^\circ$ prevailing wind direction. (a, b) Initial full-field predictions from M1; (c, d) refined predictions from M2; (e, f) corresponding LES ground truth. (g, h) Mean difference variation during the Gauss-Seidel iteration of the two cases, respectively. (i, j) Probability density functions (PDFs) of wind speed in the fluid region for the two cases. The wind speeds are shown in physical units.}
    \label{fig:Fig8}
\end{figure*}

Having examined the performance of both models individually, this section presents the full-field prediction as a cooperative framework between M1 and M2, following the procedure in \S~\ref{sec: two-stage framework}.

Figure~\ref{fig:Fig8} shows the full-field prediction performance of the proposed framework for two unseen cases, AU-Syd-U1 and CA-Van- (see \citet{Nazarian2025} for naming rules), with wind directions of $\theta = 0^\circ$ and $90^\circ$, respectively. Figures~\ref{fig:Fig8}(a,b) show the initial predictions from M1, while the corresponding ground-truth fields are shown in Figs.~\ref{fig:Fig8}(e,f). The initial estimated velocity fields generally reproduce the bulk spatial distribution of the flow: lower velocities are predicted in densely built regions, while higher velocities appear in open channels. Local acceleration and wake structures around building corners are also reasonably captured by the initial M1 estimates. However, clear discontinuities remain visible, corresponding to the boundaries of the independently predicted patches.

Figures~\ref{fig:Fig8}(c,d) show the final predictions after refinement by M2 through Gauss-Seidel iterations. The evolution of the mean absolute difference $D$ during the Gauss-Seidel iterations is shown in Figures~\ref{fig:Fig8}(g,h), which shows that the differences continuously decrease during the iterations, reach a minimum at around 25 steps in both cases, and do not decrease further over the next three steps. This is regarded as convergence.

Compared with the initial M1 fields, the patch-boundary discontinuities are largely removed in the final predictions, as intended by the inpainting-based refinement. The overall flow structure is preserved, while local velocities around building corners and channel regions are slightly enhanced. Compared with the ground truth, the coefficients of determination are $R^2 = 0.66$ and $0.69$ for the two cases, respectively, indicating an overall good agreement. The largest discrepancies mainly occur near building corners and within high-speed channel regions. 
As expected, M2 mainly modifies local details in the initial M1 predictions. However, for larger-scale structures that are less accurately estimated by M1, M2 provides limited correction. For example, the high-velocity wake marked in Fig.~\ref{fig:Fig8}(a) with the dashed rectangle is absent from the ground truth. M2 preserves this feature and makes the pattern more continuous, but it does not identify that this region should instead correspond to low velocity.

The probability density functions of the wind speed for M2 and the ground truth are shown in Figs.~\ref{fig:Fig8}(i, j). For both cases, the predicted and ground-truth distributions show substantial overlap, indicating that the overall velocity distribution is well represented. Nevertheless, the predicted fields exhibit a higher density at larger wind speeds and a lower density at smaller wind speeds compared with the ground truth. This suggests that, for these two cases, the framework tends to overestimate the mean wind speed. 
It should be noted that building areas are excluded from the PDF plots. In the training dataset, building areas are assigned zero wind speed when patches are generated. As shown in Fig.~\ref{fig:Fig8}, the framework is able to identify the building regions accurately. Nevertheless, in post-processing, the building mask can be directly obtained from the input topography and applied to the predicted field. 

\begin{figure*}
    \centering
    \includegraphics[width=17cm]{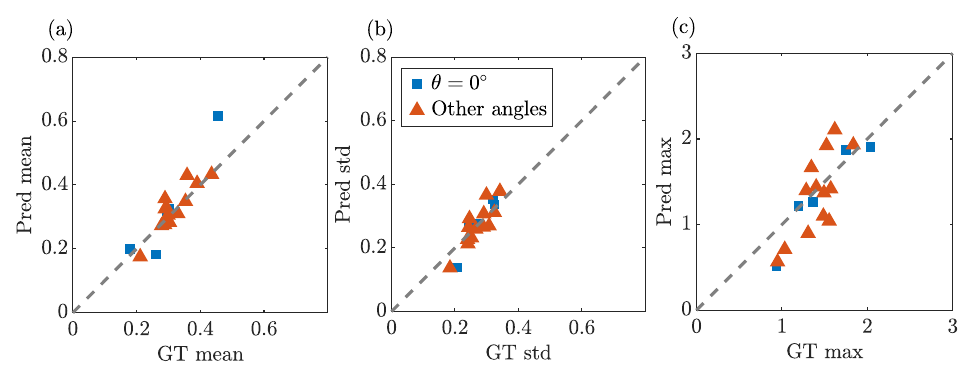}
    \caption{Scatter plots of statistics showing the performance of the entire framework in prediction. (a) Mean, (b) standard deviation, and (c) maximum values of full-field predictions over 18 unseen cases, compared with the ground truth. Cases with $0^\circ$ wind direction are shown in blue, while cases with non-zero wind directions are shown in red. The dashed line indicates perfect agreement. All the statistics are taken from the fluid region.}
    \label{fig:Fig9}
\end{figure*}

To evaluate the overall performance across all 18 unseen cases, Fig.~\ref{fig:Fig9} compares statistics of the final full-field predictions with those of the ground truth. The cases are classified into zero and non-zero wind directions, since non-zero wind directions require an additional rotation step during preprocessing.

Figure~\ref{fig:Fig9}(a) shows that the framework generally predicts the mean velocity well, with most cases lying close to the perfect-agreement line. A closer inspection shows that only two cases are noticeably underestimated, with ground-truth mean velocities of approximately $0.21$ and $0.26~\mathrm{m/s}$, respectively. The remaining cases are either close to the ground truth or slightly overestimated, which is consistent with the probability density distributions shown for the two representative cases in Fig.~\ref{fig:Fig8}. One case, with a ground-truth mean velocity of approximately $0.46~\mathrm{m/s}$, appears as a clear outlier and is substantially overestimated.
The standard deviation shows the best agreement with the ground truth, as all cases are closely distributed around the perfect-agreement line in Fig.~\ref{fig:Fig9}(b). This indicates that the relative variation of wind speed within each case is generally well preserved by the framework. 
The largest discrepancies are observed in the maximum velocity, shown in Fig.~\ref{fig:Fig9}(c), which reflects the prediction of extreme local wind speeds. Compared with the mean and standard deviation, the maximum-velocity points are more widely scattered around the perfect-agreement line. Approximately half of the cases underestimate the maximum velocity, while the remaining cases overestimate it. This suggests that although the framework captures the overall velocity pattern well, larger local discrepancies can still occur, particularly in regions associated with extreme wind speeds. This corresponds to the observation in Fig.~\ref{fig:Fig8} near building corners and within high-speed channel regions.

Finally, the statistics do not show a strong dependence on wind direction. This suggests that the rotation operation used in preprocessing for non-zero wind directions does not introduce a significant systematic effect on the full-field prediction results.

\section{Conclusions} \label{sec: conclusions}

This study developed a two-stage U-Net framework for efficient and seamless prediction of time-averaged pedestrian-level wind speed over realistic urban morphologies. The main objective was to address a key limitation of patch-wise inference: although it allows predictions over domains of arbitrary size, independently predicted patches can introduce discontinuities at patch boundaries.

The proposed framework first uses model M1 to generate an initial full-field prediction from urban morphology and upstream wind conditions. A second model, M2, then refines this field using an inpainting strategy, where a larger contextual window is used to modify the central patch based on neighbouring flow information. During full-field prediction, M2 is applied iteratively using a Gauss-Seidel scheme, so that updated patches immediately inform subsequent refinements.

The results show that M1 captures the main spatial structure of pedestrian-level wind speed and performs well for most low- and moderate-velocity regions. However, it tends to underestimate high-velocity regions, especially near building corners. This is likely related to the limited number of high-velocity samples and the smoothing effect of the loss function. M2 improves the spatial consistency of the prediction by reducing visible patch-boundary artefacts and producing more coherent full-field wind structures.

For unseen full-field cases, the framework reproduces the overall wind-speed distribution reasonably well. The mean velocity and standard deviation are generally close to the LES ground truth, indicating that the relative spatial variability is largely preserved. However, maximum velocities are slightly less accurate, showing that the current model is less reliable for extreme wind events.

The main contribution of this work is the introduction of an inpainting-based refinement model for patch-wise full-field pedestrian wind prediction. Compared with simple smoothing or overlapping-window averaging, M2 incorporates surrounding flow information directly into the prediction process. The resulting two-stage framework retains the flexibility of patch-based inference while producing more continuous full-field predictions. Future work should focus on improving the prediction of high-velocity events, for example, through reweighted loss functions, targeted sampling, or additional physical constraints.

\section*{Acknowledgements}
The authors gratefully acknowledge the support of the European Union's Horizon Europe Research and Innovation Programme through the project UrbanAIR (grant agreement No. 101188131), the EPSRC AI-Respire project (grant number EP/Y018680/1) and the ARCHER2 UK National Supercomputing Service (project ARCHER2-eCSE05-3).

{\scriptsize
\bibliographystyle{elsarticle-num-names} 
\bibliography{References.bib}

\begin{thebibliography}{41}
\expandafter\ifx\csname natexlab\endcsname\relax\def\natexlab#1{#1}\fi
\providecommand{\url}[1]{\texttt{#1}}
\providecommand{\href}[2]{#2}
\providecommand{\path}[1]{#1}
\providecommand{\DOIprefix}{doi:}
\providecommand{\ArXivprefix}{arXiv:}
\providecommand{\URLprefix}{URL: }
\providecommand{\Pubmedprefix}{pmid:}
\providecommand{\doi}[1]{\href{http://dx.doi.org/#1}{\path{#1}}}
\providecommand{\Pubmed}[1]{\href{pmid:#1}{\path{#1}}}
\providecommand{\bibinfo}[2]{#2}
\ifx\xfnm\relax \def\xfnm[#1]{\unskip,\space#1}\fi
\bibitem[{Blocken et~al.(2016)Blocken, Stathopoulos, and van Beeck}]{Blocken2016}
\bibinfo{author}{B.~Blocken}, \bibinfo{author}{T.~Stathopoulos}, \bibinfo{author}{J.~P. A.~J. van Beeck},
\newblock \bibinfo{title}{Pedestrian-level wind conditions around buildings: Review of wind-tunnel and {CFD} techniques and their accuracy for wind comfort assessment},
\newblock \bibinfo{journal}{Building and Environment} \bibinfo{volume}{100} (\bibinfo{year}{2016}) \bibinfo{pages}{50--81}.
\bibitem[{Tominaga and Stathopoulos(2011)}]{Tominaga2011}
\bibinfo{author}{Y.~Tominaga}, \bibinfo{author}{T.~Stathopoulos},
\newblock \bibinfo{title}{{CFD} modeling of pollution dispersion in a street canyon: Comparison between {LES} and {RANS}},
\newblock \bibinfo{journal}{Journal of Wind Engineering and Industrial Aerodynamics} \bibinfo{volume}{99} (\bibinfo{year}{2011}) \bibinfo{pages}{340--348}.
\bibitem[{Blocken and Carmeliet(2004)}]{Blocken2004}
\bibinfo{author}{B.~Blocken}, \bibinfo{author}{J.~Carmeliet},
\newblock \bibinfo{title}{Pedestrian wind environment around buildings: Literature review and practical examples},
\newblock \bibinfo{journal}{Journal of Thermal Envelope and Building Science} \bibinfo{volume}{28} (\bibinfo{year}{2004}) \bibinfo{pages}{107--159}.
\bibitem[{Stathopoulos(2006)}]{Stathopoulos2006}
\bibinfo{author}{T.~Stathopoulos},
\newblock \bibinfo{title}{Pedestrian level winds and outdoor human comfort},
\newblock \bibinfo{journal}{Journal of Wind Engineering and Industrial Aerodynamics} \bibinfo{volume}{94} (\bibinfo{year}{2006}) \bibinfo{pages}{769--780}.
\bibitem[{He et~al.(2019)He, Tablada, and Wong}]{He2019}
\bibinfo{author}{Y.~He}, \bibinfo{author}{A.~Tablada}, \bibinfo{author}{N.~H. Wong},
\newblock \bibinfo{title}{Effects of non-uniform and orthogonal breezeway networks on pedestrian ventilation in {S}ingapore's high-density residential neighbourhoods},
\newblock \bibinfo{journal}{Urban Climate} \bibinfo{volume}{27} (\bibinfo{year}{2019}) \bibinfo{pages}{597--614}.
\bibitem[{Lean et~al.(2024)Lean, Theeuwes, Baldauf, Barkmeijer, Bessardon, Blunn, Bojarova, Boutle, Clark, Demuzere, Dueben, Frogner, de~Haan, Harrison, Heerwaarden, Honnert, Lock, Marsigli, Masson, Mccabe, Reeuwijk, Roberts, Siebesma, Smolíková, and Yang}]{Lean2024}
\bibinfo{author}{H.~W. Lean}, \bibinfo{author}{N.~E. Theeuwes}, \bibinfo{author}{M.~Baldauf}, \bibinfo{author}{J.~Barkmeijer}, \bibinfo{author}{G.~Bessardon}, \bibinfo{author}{L.~Blunn}, \bibinfo{author}{J.~Bojarova}, \bibinfo{author}{I.~A. Boutle}, \bibinfo{author}{P.~A. Clark}, \bibinfo{author}{M.~Demuzere}, \bibinfo{author}{P.~Dueben}, \bibinfo{author}{I.-L. Frogner}, \bibinfo{author}{S.~de~Haan}, \bibinfo{author}{D.~Harrison}, \bibinfo{author}{C.~v. Heerwaarden}, \bibinfo{author}{R.~Honnert}, \bibinfo{author}{A.~Lock}, \bibinfo{author}{C.~Marsigli}, \bibinfo{author}{V.~Masson}, \bibinfo{author}{A.~Mccabe}, \bibinfo{author}{M.~v. Reeuwijk}, \bibinfo{author}{N.~Roberts}, \bibinfo{author}{P.~Siebesma}, \bibinfo{author}{P.~Smolíková}, \bibinfo{author}{X.~Yang},
\newblock \bibinfo{title}{The hectometric modelling challenge: Gaps in the current state of the art and ways forward towards the implementation of 100-m scale weather and climate models},
\newblock \bibinfo{journal}{Quarterly Journal of the Royal Meteorological Society} \bibinfo{volume}{150} (\bibinfo{year}{2024}) \bibinfo{pages}{4671--4708}.
\bibitem[{Willemsen and Wisse(2007)}]{Willemsen2007}
\bibinfo{author}{E.~Willemsen}, \bibinfo{author}{J.~A. Wisse},
\newblock \bibinfo{title}{Design for wind comfort in {T}he {N}etherlands: Procedures, criteria and open research issues},
\newblock \bibinfo{journal}{Journal of Wind Engineering and Industrial Aerodynamics} \bibinfo{volume}{95} (\bibinfo{year}{2007}) \bibinfo{pages}{1541--1550}.
\bibitem[{Kastner and Dogan(2023)}]{Kastner2023}
\bibinfo{author}{P.~Kastner}, \bibinfo{author}{T.~Dogan},
\newblock \bibinfo{title}{A {GAN}-based surrogate model for instantaneous urban wind flow prediction},
\newblock \bibinfo{journal}{Building and Environment} \bibinfo{volume}{242} (\bibinfo{year}{2023}) \bibinfo{pages}{110384}.
\bibitem[{G{\"u}r and Karadag(2024)}]{Gur2024}
\bibinfo{author}{M.~G{\"u}r}, \bibinfo{author}{I.~Karadag},
\newblock \bibinfo{title}{Machine learning for pedestrian-level wind comfort analysis},
\newblock \bibinfo{journal}{Buildings} \bibinfo{volume}{14} (\bibinfo{year}{2024}) \bibinfo{pages}{1845}.
\bibitem[{Wang and Ma(2025)}]{Wang2025}
\bibinfo{author}{H.~Wang}, \bibinfo{author}{W.~Ma},
\newblock \bibinfo{title}{Evaluating a deep learning-based surrogate model for predicting wind distribution in urban microclimate design},
\newblock \bibinfo{journal}{Building and Environment}  (\bibinfo{year}{2025}) \bibinfo{pages}{112426}.
\bibitem[{Caron et~al.(2025)Caron, Lauret, and Bastide}]{Caron2025}
\bibinfo{author}{C.~Caron}, \bibinfo{author}{P.~Lauret}, \bibinfo{author}{A.~Bastide},
\newblock \bibinfo{title}{Machine learning to speed up computational fluid dynamics engineering simulations for built environments: A review},
\newblock \bibinfo{journal}{Building and Environment} \bibinfo{volume}{267} (\bibinfo{year}{2025}) \bibinfo{pages}{112229}.
\bibitem[{Xiang et~al.(2021)Xiang, Fu, Zhou, Wang, Zhang, Hu, Xu, Liu, Liu, Ma, and Tao}]{Xiang2021}
\bibinfo{author}{S.~Xiang}, \bibinfo{author}{X.~Fu}, \bibinfo{author}{J.~Zhou}, \bibinfo{author}{Y.~Wang}, \bibinfo{author}{Y.~Zhang}, \bibinfo{author}{X.~Hu}, \bibinfo{author}{J.~Xu}, \bibinfo{author}{H.~Liu}, \bibinfo{author}{J.~Liu}, \bibinfo{author}{J.~Ma}, \bibinfo{author}{S.~Tao},
\newblock \bibinfo{title}{{Non-intrusive reduced order model of urban airflow with dynamic boundary conditions}},
\newblock \bibinfo{journal}{Building and Environment} \bibinfo{volume}{187} (\bibinfo{year}{2021}) \bibinfo{pages}{107397}.
\bibitem[{Masoumi-Verki et~al.(2022)Masoumi-Verki, Haghighat, and Eicker}]{Masoumi-Verki2022}
\bibinfo{author}{S.~Masoumi-Verki}, \bibinfo{author}{F.~Haghighat}, \bibinfo{author}{U.~Eicker},
\newblock \bibinfo{title}{{Improving the performance of a CAE-based reduced-order model for predicting turbulent airflow field around an isolated high-rise building}},
\newblock \bibinfo{journal}{Sustainable Cities and Society} \bibinfo{volume}{87} (\bibinfo{year}{2022}) \bibinfo{pages}{104252}.
\bibitem[{Quilodrán-Casas and Arcucci(2023)}]{Quilodran-Casas2023}
\bibinfo{author}{C.~Quilodrán-Casas}, \bibinfo{author}{R.~Arcucci},
\newblock \bibinfo{title}{{A data-driven adversarial machine learning for 3D surrogates of unstructured computational fluid dynamic simulations}},
\newblock \bibinfo{journal}{Physica A: Statistical Mechanics and its Applications} \bibinfo{volume}{615} (\bibinfo{year}{2023}) \bibinfo{pages}{128564}.
\bibitem[{Ronneberger et~al.(2015)Ronneberger, Fischer, and Brox}]{Unet2015}
\bibinfo{author}{O.~Ronneberger}, \bibinfo{author}{P.~Fischer}, \bibinfo{author}{T.~Brox},
\newblock \bibinfo{title}{U-net: Convolutional networks for biomedical image segmentation},
\newblock in: \bibinfo{editor}{N.~Navab}, \bibinfo{editor}{J.~Hornegger}, \bibinfo{editor}{W.~M. Wells}, \bibinfo{editor}{A.~F. Frangi} (Eds.), \bibinfo{booktitle}{Medical Image Computing and Computer-Assisted Intervention -- MICCAI 2015}, \bibinfo{publisher}{Springer International Publishing}, \bibinfo{address}{Cham}, \bibinfo{year}{2015}, pp. \bibinfo{pages}{234--241}.
\bibitem[{Lu et~al.(2023)Lu, Zhou, Xiao, and Li}]{Lu2022}
\bibinfo{author}{Y.~Lu}, \bibinfo{author}{X.-H. Zhou}, \bibinfo{author}{H.~Xiao}, \bibinfo{author}{Q.~Li},
\newblock \bibinfo{title}{Using machine learning to predict urban canopy flows for land surface modeling},
\newblock \bibinfo{journal}{Geophysical Research Letters} \bibinfo{volume}{50} (\bibinfo{year}{2023}) \bibinfo{pages}{e2022GL102313}.
\bibitem[{Clarke et~al.(2024)Clarke, Giljarhus, Oggiano, Saddington, and Depuru-Mohan}]{Clarke2024}
\bibinfo{author}{A.~Clarke}, \bibinfo{author}{K.~E.~T. Giljarhus}, \bibinfo{author}{L.~Oggiano}, \bibinfo{author}{A.~Saddington}, \bibinfo{author}{K.~Depuru-Mohan},
\newblock \bibinfo{title}{{MLP}-mixer-based deep learning network for pedestrian-level wind assessment},
\newblock \bibinfo{journal}{Environmental Data Science} \bibinfo{volume}{3} (\bibinfo{year}{2024}) \bibinfo{pages}{e35}.
\bibitem[{Cui et~al.(2025)Cui, Li, and Shen}]{Cui2025}
\bibinfo{author}{X.~Cui}, \bibinfo{author}{Y.~Li}, \bibinfo{author}{P.~Shen},
\newblock \bibinfo{title}{Beyond {CFD}: explainable machine learning for efficient assessment of urban morphology impacts on pedestrian level wind and thermal environment},
\newblock \bibinfo{journal}{Journal of Building Performance Simulation}  (\bibinfo{year}{2025}) \bibinfo{pages}{1--16}.
\bibitem[{Briegel et~al.(2025)Briegel, Schrodi, Sulzer, Brox, Pinto, and Christen}]{Briegel2025}
\bibinfo{author}{F.~Briegel}, \bibinfo{author}{S.~Schrodi}, \bibinfo{author}{M.~Sulzer}, \bibinfo{author}{T.~Brox}, \bibinfo{author}{J.~G. Pinto}, \bibinfo{author}{A.~Christen},
\newblock \bibinfo{title}{Deep learning enables city-wide climate projections of street-level heat stress},
\newblock \bibinfo{journal}{Urban Climate} \bibinfo{volume}{62} (\bibinfo{year}{2025}) \bibinfo{pages}{102564}.
\bibitem[{Lu et~al.(2025)Lu, Li, Hobeichi, Azad, and Nazarian}]{Lu2025}
\bibinfo{author}{J.~Lu}, \bibinfo{author}{W.~Li}, \bibinfo{author}{S.~Hobeichi}, \bibinfo{author}{S.~A. Azad}, \bibinfo{author}{N.~Nazarian},
\newblock \bibinfo{title}{Machine learning predicts pedestrian wind flow from urban morphology and prevailing wind direction},
\newblock \bibinfo{journal}{Environmental Research Letters} \bibinfo{volume}{20} (\bibinfo{year}{2025}) \bibinfo{pages}{054006}.
\bibitem[{Vargiemezis and Gorlé(2025)}]{Vargiemezis2025}
\bibinfo{author}{T.~Vargiemezis}, \bibinfo{author}{C.~Gorlé},
\newblock \bibinfo{title}{From large-eddy simulations to deep learning: A {U}-net model for fast urban canopy flow predictions},
\newblock \bibinfo{journal}{Sustainable Cities and Society} \bibinfo{volume}{135} (\bibinfo{year}{2025}) \bibinfo{pages}{107005}.
\bibitem[{Snaiki et~al.(2026)Snaiki, Lu, Li, and Nazarian}]{Snaiki2026}
\bibinfo{author}{R.~Snaiki}, \bibinfo{author}{J.~Lu}, \bibinfo{author}{S.~Li}, \bibinfo{author}{N.~Nazarian},
\newblock \bibinfo{title}{A hierarchical deep learning model for predicting pedestrian-level urban winds},
\newblock \bibinfo{journal}{Building and Environment} \bibinfo{volume}{294} (\bibinfo{year}{2026}) \bibinfo{pages}{114354}.
\bibitem[{Maggiori et~al.(2017)Maggiori, Tarabalka, Charpiat, and Alliez}]{Maggiori2017}
\bibinfo{author}{E.~Maggiori}, \bibinfo{author}{Y.~Tarabalka}, \bibinfo{author}{G.~Charpiat}, \bibinfo{author}{P.~Alliez},
\newblock \bibinfo{title}{Convolutional neural networks for large-scale remote-sensing image classification},
\newblock \bibinfo{journal}{IEEE Transactions on Geoscience and Remote Sensing} \bibinfo{volume}{55} (\bibinfo{year}{2017}) \bibinfo{pages}{645--657}.
\bibitem[{Volpi and Tuia(2017)}]{Volpi2017}
\bibinfo{author}{M.~Volpi}, \bibinfo{author}{D.~Tuia},
\newblock \bibinfo{title}{Dense semantic labeling of subdecimeter resolution images with convolutional neural networks},
\newblock in: \bibinfo{booktitle}{IEEE Transactions on Geoscience and Remote Sensing}, volume~\bibinfo{volume}{55}, \bibinfo{year}{2017}, pp. \bibinfo{pages}{881--893}.
\bibitem[{Clemente et~al.(2023)Clemente, Giljarhus, Oggiano, and Ruocco}]{Clemente2023}
\bibinfo{author}{A.~V. Clemente}, \bibinfo{author}{K.~E.~T. Giljarhus}, \bibinfo{author}{L.~Oggiano}, \bibinfo{author}{M.~Ruocco}, \bibinfo{title}{Configurable convolutional neural networks for real-time pedestrian-level wind prediction in urban environments}, \bibinfo{year}{2023}. \href{http://arxiv.org/abs/2311.07985}{{\tt arXiv:2311.07985}}.
\bibitem[{Calafell et~al.(2026)Calafell, Bustillo, Armengol, Gómez, Ramirez-Javega, and Lehmkuhl}]{Calafell2026}
\bibinfo{author}{J.~Calafell}, \bibinfo{author}{J.~Bustillo}, \bibinfo{author}{J.~Armengol}, \bibinfo{author}{S.~Gómez}, \bibinfo{author}{F.~Ramirez-Javega}, \bibinfo{author}{O.~Lehmkuhl},
\newblock \bibinfo{title}{Building a general and data-efficient convolutional neural network-based model for fast urban flow estimation},
\newblock \bibinfo{journal}{Building and Environment} \bibinfo{volume}{296} (\bibinfo{year}{2026}) \bibinfo{pages}{114488}.
\bibitem[{Reina et~al.(2020)Reina, Panchumarthy, Thakur, Bastidas, and Bakas}]{Reina2020}
\bibinfo{author}{G.~A. Reina}, \bibinfo{author}{R.~Panchumarthy}, \bibinfo{author}{S.~P. Thakur}, \bibinfo{author}{A.~Bastidas}, \bibinfo{author}{S.~Bakas},
\newblock \bibinfo{title}{Systematic evaluation of image tiling adverse effects on deep learning semantic segmentation},
\newblock \bibinfo{journal}{Frontiers in Neuroscience}  (\bibinfo{year}{2020}).
\bibitem[{Innamorati et~al.(2020)Innamorati, Ritschel, Weyrich, and Mitra}]{Innamorati2020}
\bibinfo{author}{C.~Innamorati}, \bibinfo{author}{T.~Ritschel}, \bibinfo{author}{T.~Weyrich}, \bibinfo{author}{N.~J. Mitra},
\newblock \bibinfo{title}{Learning on the edge: Investigating boundary filters in {CNN}s},
\newblock \bibinfo{journal}{International Journal of Computer Vision} \bibinfo{volume}{128} (\bibinfo{year}{2020}) \bibinfo{pages}{773--782}.
\bibitem[{Isensee et~al.(2019)Isensee, Petersen, Klein, Zimmerer, Jaeger, Kohl, Wasserthal, Koehler, Norajitra, Wirkert, and Maier-Hein}]{Isensee2018}
\bibinfo{author}{F.~Isensee}, \bibinfo{author}{J.~Petersen}, \bibinfo{author}{A.~Klein}, \bibinfo{author}{D.~Zimmerer}, \bibinfo{author}{P.~F. Jaeger}, \bibinfo{author}{S.~Kohl}, \bibinfo{author}{J.~Wasserthal}, \bibinfo{author}{G.~Koehler}, \bibinfo{author}{T.~Norajitra}, \bibinfo{author}{S.~Wirkert}, \bibinfo{author}{K.~H. Maier-Hein},
\newblock \bibinfo{title}{Abstract: {nnU-Net}: Self-adapting framework for u-net-based medical image segmentation},
\newblock in: \bibinfo{editor}{H.~Handels}, \bibinfo{editor}{T.~M. Deserno}, \bibinfo{editor}{A.~Maier}, \bibinfo{editor}{K.~H. Maier-Hein}, \bibinfo{editor}{C.~Palm}, \bibinfo{editor}{T.~Tolxdorff} (Eds.), \bibinfo{booktitle}{Bildverarbeitung f{\"u}r die Medizin 2019}, \bibinfo{publisher}{Springer Fachmedien Wiesbaden}, \bibinfo{address}{Wiesbaden}, \bibinfo{year}{2019}, pp. \bibinfo{pages}{22--22}.
\bibitem[{Pielawski and W{\"a}hlby(2020)}]{Pielawski2020}
\bibinfo{author}{N.~Pielawski}, \bibinfo{author}{C.~W{\"a}hlby},
\newblock \bibinfo{title}{Introducing {Hann} windows for reducing edge-effects in patch-based image segmentation},
\newblock \bibinfo{journal}{PLOS ONE} \bibinfo{volume}{15} (\bibinfo{year}{2020}) \bibinfo{pages}{e0229839}.
\bibitem[{Abdellatif et~al.(2024)Abdellatif, Elsheikh, and Menke}]{Abdellatif2024}
\bibinfo{author}{A.~Abdellatif}, \bibinfo{author}{A.~H. Elsheikh}, \bibinfo{author}{H.~P. Menke},
\newblock \bibinfo{title}{Local padding in patch-based {GAN}s for seamless infinite-sized texture synthesis},
\newblock \bibinfo{year}{2024}.
\bibitem[{Jeon et~al.(2025)Jeon, Yang, Fu, and Feng}]{Jeon2025}
\bibinfo{author}{Y.~S. Jeon}, \bibinfo{author}{H.~Yang}, \bibinfo{author}{H.~Fu}, \bibinfo{author}{M.~Feng}, \bibinfo{title}{No more sliding window: Efficient {3D} medical image segmentation with differentiable top-k patch sampling}, \bibinfo{year}{2025}. \href{http://arxiv.org/abs/2501.10814}{{\tt arXiv:2501.10814}}.
\bibitem[{Heaney et~al.(2022)Heaney, Liu, Go, Wolffs, Salinas, Navon, and Pain}]{Heaney2022}
\bibinfo{author}{C.~E. Heaney}, \bibinfo{author}{X.~Liu}, \bibinfo{author}{H.~Go}, \bibinfo{author}{Z.~Wolffs}, \bibinfo{author}{P.~Salinas}, \bibinfo{author}{I.~M. Navon}, \bibinfo{author}{C.~C. Pain},
\newblock \bibinfo{title}{Extending the capabilities of data-driven reduced-order models to make predictions for unseen scenarios: applied to flow around buildings},
\newblock \bibinfo{journal}{Frontiers in Physics} \bibinfo{volume}{10} (\bibinfo{year}{2022}) \bibinfo{pages}{910381}.
\bibitem[{Pathak et~al.(2016)Pathak, Kr\"ahenb\"uhl, Donahue, Darrell, and Efros}]{Pathak2016}
\bibinfo{author}{D.~Pathak}, \bibinfo{author}{P.~Kr\"ahenb\"uhl}, \bibinfo{author}{J.~Donahue}, \bibinfo{author}{T.~Darrell}, \bibinfo{author}{A.~A. Efros},
\newblock \bibinfo{title}{Context encoders: Feature learning by inpainting},
\newblock in: \bibinfo{booktitle}{Proceedings of the IEEE Conference on Computer Vision and Pattern Recognition (CVPR)}, \bibinfo{year}{2016}, pp. \bibinfo{pages}{2536--2544}.
\bibitem[{Liu et~al.(2018)Liu, Reda, Shih, Wang, Tao, and Catanzaro}]{Liu2018}
\bibinfo{author}{G.~Liu}, \bibinfo{author}{F.~A. Reda}, \bibinfo{author}{K.~J. Shih}, \bibinfo{author}{T.-C. Wang}, \bibinfo{author}{A.~Tao}, \bibinfo{author}{B.~Catanzaro},
\newblock \bibinfo{title}{Image inpainting for irregular holes using partial convolutions},
\newblock in: \bibinfo{booktitle}{Proceedings of the European Conference on Computer Vision (ECCV)}, \bibinfo{year}{2018}, pp. \bibinfo{pages}{85--100}.
\bibitem[{Yu et~al.(2018)Yu, Lin, Yang, Shen, Lu, and Huang}]{Yu2018_inpainting}
\bibinfo{author}{J.~Yu}, \bibinfo{author}{Z.~Lin}, \bibinfo{author}{J.~Yang}, \bibinfo{author}{X.~Shen}, \bibinfo{author}{X.~Lu}, \bibinfo{author}{T.~S. Huang},
\newblock \bibinfo{title}{Generative image inpainting with contextual attention},
\newblock in: \bibinfo{booktitle}{Proceedings of the IEEE Conference on Computer Vision and Pattern Recognition (CVPR)}, \bibinfo{year}{2018}, pp. \bibinfo{pages}{5505--5514}.
\bibitem[{Nazarian et~al.(2025)Nazarian, Lu, Lipson, Hart, Liu, Krayenhoff, Blunn, and Martilli}]{Nazarian2025}
\bibinfo{author}{N.~Nazarian}, \bibinfo{author}{J.~Lu}, \bibinfo{author}{M.~J. Lipson}, \bibinfo{author}{M.~A. Hart}, \bibinfo{author}{S.~Liu}, \bibinfo{author}{E.~S. Krayenhoff}, \bibinfo{author}{L.~Blunn}, \bibinfo{author}{A.~Martilli},
\newblock \bibinfo{title}{{UrbanTALES}: A large-eddy simulation dataset for urban canopy layer turbulence and parameterization},
\newblock \bibinfo{journal}{Bulletin of the American Meteorological Society} \bibinfo{volume}{106} (\bibinfo{year}{2025}) \bibinfo{pages}{E2461 -- E2478}.
\bibitem[{Maronga et~al.(2015)Maronga, Gryschka, Heinze, Hoffmann, Kanani-S\"uhring, Keck, Ketelsen, Letzel, S\"uhring, and Raasch}]{PALM2015}
\bibinfo{author}{B.~Maronga}, \bibinfo{author}{M.~Gryschka}, \bibinfo{author}{R.~Heinze}, \bibinfo{author}{F.~Hoffmann}, \bibinfo{author}{F.~Kanani-S\"uhring}, \bibinfo{author}{M.~Keck}, \bibinfo{author}{K.~Ketelsen}, \bibinfo{author}{M.~O. Letzel}, \bibinfo{author}{M.~S\"uhring}, \bibinfo{author}{S.~Raasch},
\newblock \bibinfo{title}{The parallelized large-eddy simulation model ({PALM}) version 4.0 for atmospheric and oceanic flows: model formulation, recent developments, and future perspectives},
\newblock \bibinfo{journal}{Geoscientific Model Development} \bibinfo{volume}{8} (\bibinfo{year}{2015}) \bibinfo{pages}{2515--2551}.
\bibitem[{Holmes(2015)}]{Holmes2015}
\bibinfo{author}{J.~D. Holmes}, \bibinfo{title}{Wind Loading of Structures}, \bibinfo{edition}{3rd} ed., \bibinfo{publisher}{CRC Press}, \bibinfo{address}{Boca Raton}, \bibinfo{year}{2015}.
\bibitem[{Stull(1988)}]{Stull1988}
\bibinfo{author}{R.~B. Stull}, \bibinfo{title}{An Introduction to Boundary Layer Meteorology}, \bibinfo{publisher}{Springer}, \bibinfo{year}{1988}.
\bibitem[{Wang et~al.(2004)Wang, Bovik, Sheikh, and Simoncelli}]{Wang2004}
\bibinfo{author}{Z.~Wang}, \bibinfo{author}{A.~C. Bovik}, \bibinfo{author}{H.~R. Sheikh}, \bibinfo{author}{E.~P. Simoncelli},
\newblock \bibinfo{title}{Image quality assessment: from error visibility to structural similarity},
\newblock \bibinfo{journal}{IEEE Transactions on Image Processing} \bibinfo{volume}{13} (\bibinfo{year}{2004}) \bibinfo{pages}{600--612}.

\end{thebibliography}
}

\end{document}